\documentclass[journal]{IEEEtran}
\usepackage{latexsym}

\usepackage{enumerate}
\usepackage{graphicx}
\usepackage{subfigure}
\usepackage{multirow}

\usepackage{framed,multirow}
\usepackage{times}
\usepackage{graphicx} 
\usepackage{subfigure}
\usepackage{amssymb}
\usepackage{latexsym}

\usepackage{algorithm}
\usepackage{algorithmic}

\usepackage{hyperref}

\usepackage{url}
\usepackage{xcolor}
\usepackage{amsmath}
\usepackage{amsxtra}
\usepackage{braket}
\usepackage{array}
\usepackage{color}
\usepackage{url}
\usepackage{placeins}
\usepackage{threeparttable}

\usepackage{times}
\usepackage{balance}
\usepackage{enumerate}
\usepackage{multirow}
\usepackage{pifont}
\usepackage{marvosym}
\usepackage{ifsym}

\begin{document}
\title{Graph Convolutional Neural Networks based on Quantum Vertex Saliency}
\author{Lu~Bai,~\IEEEmembership{}Yuhang~Jiao,~\IEEEmembership{}Luca~Rossi,~\IEEEmembership{}Lixin~Cui,~\IEEEmembership{}Jian Cheng,~\IEEEmembership{}Edwin~R.~Hancock,~\IEEEmembership{IAPR~Fellow,~IEEE~Fellow}

\thanks{Lu Bai is with Central University of Finance and Economics, Beijing, China. e-mail: bailucs@cufe.edu.cn; bailu69@hotmail.com.}
\thanks{Lixin Cui and Yuhang Jiao is with Central University of Finance and Economics, Beijing, China.}
\thanks{Luca Rossi is with Southern University of Science and Technology, Shenzhen, China.}
\thanks{Jian Cheng is with National Laboratory of Pattern Recognition, Institute of Automation, Chinese Academy of Sciences, Beijing, China.}
\thanks{Edwin R. Hancock is with University of York, York, UK. }}

\markboth{Journal of \LaTeX\ Class Files,~Vol.~6, No.~1, January~2007}%
{Shell \MakeLowercase{\textit{et al.}}: Bare Demo of IEEEtran.cls
for Journals}
\maketitle

\begin{abstract}
This paper proposes a new Quantum Spatial Graph Convolutional Neural Network (QSGCNN) model that can directly learn a classification function for graphs of arbitrary sizes. Unlike state-of-the-art Graph Convolutional Neural Network (GCNN) models, the proposed QSGCNN model incorporates the process of identifying transitive aligned vertices between graphs, and transforms arbitrary sized graphs into fixed-sized aligned vertex grid structures. In order to learn representative graph characteristics, a new quantum spatial graph convolution is proposed and employed to extract multi-scale vertex features, in terms of quantum information propagation between grid vertices of each graph. Since the quantum spatial convolution preserves the grid structures of the input vertices (i.e., the convolution layer does not change the original spatial sequence of vertices), the proposed QSGCNN model allows to directly employ the traditional convolutional neural network architecture to further learn from the global graph topology, providing an end-to-end deep learning architecture that integrates the graph representation and learning in the quantum spatial graph convolution layer and the traditional convolutional layer for graph classifications. We demonstrate the effectiveness of the proposed QSGCNN model in relation to existing state-of-the-art methods. The proposed QSGCNN model addresses the shortcomings of information loss and imprecise information representation arising in existing GCN models associated with the use of SortPooling or SumPooling layers. Experiments on benchmark graph classification datasets demonstrate the effectiveness of the proposed QSGCNN model.
\end{abstract}


\IEEEpeerreviewmaketitle

\section{Introduction}\label{s1}
\IEEEPARstart{G}raph-based representations have been widely employed to model and analyze data that lies on high-dimensional non-Euclidean domains and that is naturally described in terms of pairwise relationships between its parts~\cite{DBLP:conf/nips/DefferrardBV16,DBLP:journals/tnn/ZambonAL18}. Typical instances where data can be represented using graphs include a) classifying proteins or chemical compounds~\cite{DBLP:conf/icml/KriegeM12,DBLP:journals/tnn/WuPZZY18}, b) recognizing objects from digital images~\cite{DBLP:conf/cvpr/HarchaouiB07}, c) visualizing social networks~\cite{DBLP:conf/kdd/WangC016}. A fundamental challenge arising in the analysis of real-world data represented as graphs is the lack of a clear and accurate way to represent discrete graph structures as numeric features that can be directly analyzed by standard machine learning techniques~\cite{DBLP:journals/tsmc/RiesenB09}. This paper aims to develop a new graph convolutional neural network using quantum vertex saliency, for the purpose of graph classification. Our method is based on identifying the transitive alignment information between vertices of all different graphs. That is, given three vertices $v$, $w$ and $x$ from three sample graphs, suppose $v$ and $x$ are aligned, and $w$ and $x$ are aligned, the proposed model can guarantee that $v$ and $w$ are also aligned. The alignment procedure not only provides a way of mapping each graph into a fixed-sized vertex grid structure, but also bridges the gap between the graph convolution layer and the traditional convolutional neural network layer.

\subsection{Literature Review}
There have been a large number of methods aimed at converting graph structures into numeric representations, thus providing a way of directly applying standard machine learning algorithm to problems of graph classification or clustering. Generally speaking, in the last three decades, most classical state-of-the-art approaches to the analysis of graph structures can be divided into two classes, namely 1) graph embedding methods and 2) graph kernels. The methods from the first class aim to represent graphs as vectors of permutation invariant features, so that one can directly employ standard vectorial machine learning algorithms~\cite{DBLP:journals/pr/GibertVB12,DBLP:journals/pami/WilsonHL05,DBLP:conf/icml/KondorB08,DBLP:journals/tnn/RenWH11}. All of the previous approaches are based on the computation of explicit embeddings into low dimensional vector spaces, which inevitably leads to the loss of structural information. Graph kernels, on the other hand, try to soften this limitation by (implicitly) mapping graphs to a high dimensional Hilbert space where the structural information is better preserved~\cite{DBLP:series/smpai/NeuhausB07,DBLP:journals/tnn/OnetoNDRSAA18}. The majority of state-of the-art graph kernels are instances of the R-convolution kernel originally proposed by Haussler~\cite{haussler99convolution}. The main idea underpinning R-convolution kernels is that of decomposing graphs into substructures (e.g, walks, paths, subtrees, and subgraphs) and then to measure the similarity between a pair of input graphs in terms of the similarity between their constituent substructures. Representative R-convolution graph kernels include the Weisfeiler-Lehman subtree kernel~\cite{shervashidze2010weisfeiler}, the subgraph matching kernel~\cite{DBLP:conf/icml/KriegeM12}, the backtracless path kernel~\cite{DBLP:journals/tnn/AzizWH13}, the tree-based continuous attributed kernel~\cite{DBLP:journals/tnn/MartinoNS18}, and the aligned subtree kernel~\cite{DBLP:conf/icml/Bai0ZH15}. A common limitation shared by both graph embedding methods and kernels is that of ignoring information from multiple graphs. This is because graph embedding methods usually capture structural features of individual graphs, while graph kernels reflect structural characteristics for pairs of graphs.

Recently, deep learning networks have emerged as an effective way to extract highly meaningful statistical patterns in large-scale and high-dimensional data~\cite{DBLP:series/acvpr/978-3-319-42998-4}. As evidenced by their recent successes in computer vision problems, convolutional neural networks (CNNs)~\cite{DBLP:conf/cvpr/VinyalsTBE15,DBLP:journals/cacm/KrizhevskySH17} are one of the most popular class of deep learning architectures and many researchers have devoted their efforts to generalizing CNNs to the graph domain~\cite{DBLP:journals/corr/abs-1708-02218}. Unfortunately, applying CNNs for graphs in a straightforward way is not trivial, since these networks are designed to operate on regular grids~\cite{DBLP:conf/nips/DefferrardBV16} and the associated operations of convolution, pooling and weight-sharing cannot be easily extended to graphs.

To address the aforementioned problem, two popular strategies have been proposed and employed to extend convolutional neural networks to graph domains, i.e., the spectral and the spatial strategies. Specifically, approaches using the spectral strategy utilise the property of the convolution operator from the graph Fourier
domain, and relate to the graph Laplacian~\cite{DBLP:journals/corr/BrunaZSL13,DBLP:conf/nips/RippelSA15,DBLP:journals/corr/HenaffBL15}. By transforming the graph into the spectral domain through the Laplacian matrix eigenvectors, the filter operation is performed by multiplying the graph by a series of filter coefficients. Unfortunately, most spectral-based approaches demand the size of the graph structures to be the same and cannot be performed on graphs with different sizes and Fourier bases. As a result, approaches based on the spectral strategy are usually applied to vertex classification tasks. By contrast, methods based on the spatial strategy are not restricted to the same graph structure. These methods generalize the convolution operation to the spatial structure of a graph by propagating features between neighboring vertices~\cite{DBLP:journals/corr/VialatteGM16}. For instance, Duvenaud et al.~\cite{DBLP:conf/nips/DuvenaudMABHAA15} have proposed a Neural Graph Fingerprint Network by propagating vertex features between their $1$-layer neighbors to simulate the traditional circular fingerprint. Atwood and Towsley~\cite{DBLP:conf/nips/AtwoodT16} have proposed a Diffusion Convolution Neural Network by propagating vertex features between neighbors of different layers rooted at a vertex. Although spatially based approaches can be directly applied to real-world graph classification problems, most existing methods have fairly poor performance on graph classification. This is because these methods tend to directly sum up the extracted local-level vertex features from the convolution operation as global-level graph features through a SumPooling layer. It is then difficult to learn the topological information residing in a graph through these global features.

To overcome the shortcoming of the graph convolutional neural networks associated with SumPooling, unlike the works in~\cite{DBLP:conf/nips/DuvenaudMABHAA15} and~\cite{DBLP:conf/nips/AtwoodT16}, Nieper et al.~\cite{DBLP:conf/icml/NiepertAK16} have developed a different graph convolutional neural network by constructing a fixed-sized local neighborhood for each vertex and re-ordering the vertices based on graph labeling methods and graph canonization tools. This procedure naturally forms a fixed-sized vertex grid structure for each graph, and the graph convolution operation can be performed by sliding a fixed-sized filter over spatially neighboring vertices. This operation is similar to that performed on images with standard convolutional neural networks. Zhang et al.~\cite{DBLP:conf/aaai/ZhangCNC18} have developed a novel Deep Graph Convolutional Neural Network model that can preserve more vertex information and learn from the global graph topology. Specifically, this model utilizes a newly developed SortPooling layer, that can transform the extracted vertex features of unordered vertices from spatial graph convolution layers into a fixed-sized vertex grid structure. Then a traditional convolutional neural networks can be applied to the grid structures to further learn the graph topological information.

Although both methods of Nieper et al.~\cite{DBLP:conf/icml/NiepertAK16} and Zhang et al.~\cite{DBLP:conf/aaai/ZhangCNC18} outperform state-of-the-art graph convolutional neural network models and graph kernels on graph classification tasks, these approaches suffer from the drawback of ignoring structural correspondence information between graphs, or rely on simple but inaccurate heuristics to align the vertices of the graphs, i.e., they sort the vertex orders based on the local structure descriptor of each individual graph and ignore the vertex correspondence information between different graphs. As a result, both the methods cannot reflect the precise topological correspondence information for graph structures. Moreover, these approaches also lead to significant information loss. This usually occurs when these approaches form the fixed-sized vertex grid structure and some vertices associated with lower ranking may be discarded. In summary, developing effective methods to preserve the structural information residing in graphs still remains a significant challenge.

\subsection{Contribution}
The aim in this paper is to overcome the shortcomings of the aforementioned methods by developing a new spatial graph convolutional neural network model. One key innovation of the new model is the identification of the transitive vertex alignment information between graphs. Specifically, the new model can employ the transitive alignment information to map different sized graphs into fixed-sized aligned representations, i.e., it can transform different graphs into fixed-sized aligned grid structures with consistent vertex orders. Note that the aligned grid structure can precisely integrate the structural correspondence information and preserve both the original graph topology and the vertex feature information without any information loss, since all the original vertex information will be mapped into the grid structure through the transitive alignment. Thus, it not only bridges the gap between the spatial graph convolution layer and the traditional convolutional neural network layer, but also addresses the shortcomings of information loss and imprecise information representation arising in most state-of-the-art graph convolutional neural networks associated with SortPooling or SumPooling layers. Overall, the main contributions of this work are threefold.

First, we develop a new framework for transitively aligning the vertices of a family of graphs in terms of vertex point matching. This framework can establish reliable vertex correspondence information between graphs, by gradually minimizing the inner-vertex-cluster sum of squares over the vertices of all graphs. We show that this framework can be further employed to map graphs of arbitrary sizes into fixed-sized aligned vertex grid structures, integrating precise structural correspondence information and thus minimising the loss of structural information. The resulting grid structures can bridge the gap between the spatial graph convolution layer and the traditional convolutional neural network layer.

Second, with the aligned vertex grid structures and their associated adjacency matrices to hand, we propose a novel quantum spatial graph convolution layer to extract multi-scale vertex features in terms of the quantum vertex information propagation. More specifically, we use the average mixing matrix associated with continuous-time quantum walks. We show that the new convolution layer theoretically overcomes the shortcoming of popular graph convolutional neural networks and graph kernels, supporting the empirical evidence collected in our experimental validation. Moreover, since the proposed convolution layer does not change the original spatial sequence of vertices, it allows us to directly employ the traditional convolutional neural network to further learn from the global graph topology, providing an end-to-end deep learning architecture that integrates the graph representation and learning into both the quantum spatial graph convolution and the traditional convolutional layers for graph classifications.

Third, we empirically evaluate the proposed Quantum Spatial Graph Convolutional Neural Network (QSGCNN). Experimental results on benchmark graph classification datasets demonstrate that our proposed QSGCNN significantly outperforms state-of-the-art graph kernels and deep graph convolutional network models for graph classifications.


\section{Preliminary Concepts}\label{s2}


\subsection{Continuous-time Quantum Walks}\label{s2.1}
One main objective of this work is to develop a new spatial graph convolution layer to extract multi-scale vertex features by gradually propagating information for each vertex to its neighboring vertices as well as the vertex itself. This usually requires connection information between each vertex and its neighboring vertices. Most existing methods employ the vertex adjacency matrix of each graph in the formulation of the information propagation framework~\cite{DBLP:conf/nips/DuvenaudMABHAA15,DBLP:conf/nips/AtwoodT16,DBLP:conf/icml/NiepertAK16,DBLP:conf/aaai/ZhangCNC18}. In order to capture richer vertex features from the proposed graph convolutional layer, in this work we propose to employ the vertex information propagation process of the continuous-time quantum walk. This is the quantum analogue of the classical continuous-time random walk~\cite{farhi1998quantum}.

The main reason for relying on quantum walks is that, unlike classical random walks, whose state is described by a real-valued vector
and where the evolution is governed by a doubly stochastic matrix, the state
vector of the quantum walks is complex-valued and its evolution is governed by a time-varying
unitary matrix. Thus, the quantum walk evolution is reversible, implying that it is non-ergodic and does not possess a limiting distribution. As a result, the behaviour of quantum walks is significantly different from their classical counterpart and possesses a number of important properties, e.g., it allows interference to take place. This interference, in turn, helps to reduce the tottering problem of random walks, as a quantum walkers backtracking on an edge does so with opposite phase. Furthermore, since the evolution of the quantum walk is not dominated by the low frequency components of the Laplacian spectrum, it has better ability to distinguish different graph structures. In Section~\ref{s3}, we will show that the proposed graph convolutional layer associated with the continuous-time quantum can not only reduce the tottering problem arising in some state-of-the-art graph kernels and graph convolutional network models, but also better discriminates between different graph structures.

\begin{figure*}
 \vspace{-0pt}
 \centering
\includegraphics[width=1.0\linewidth]{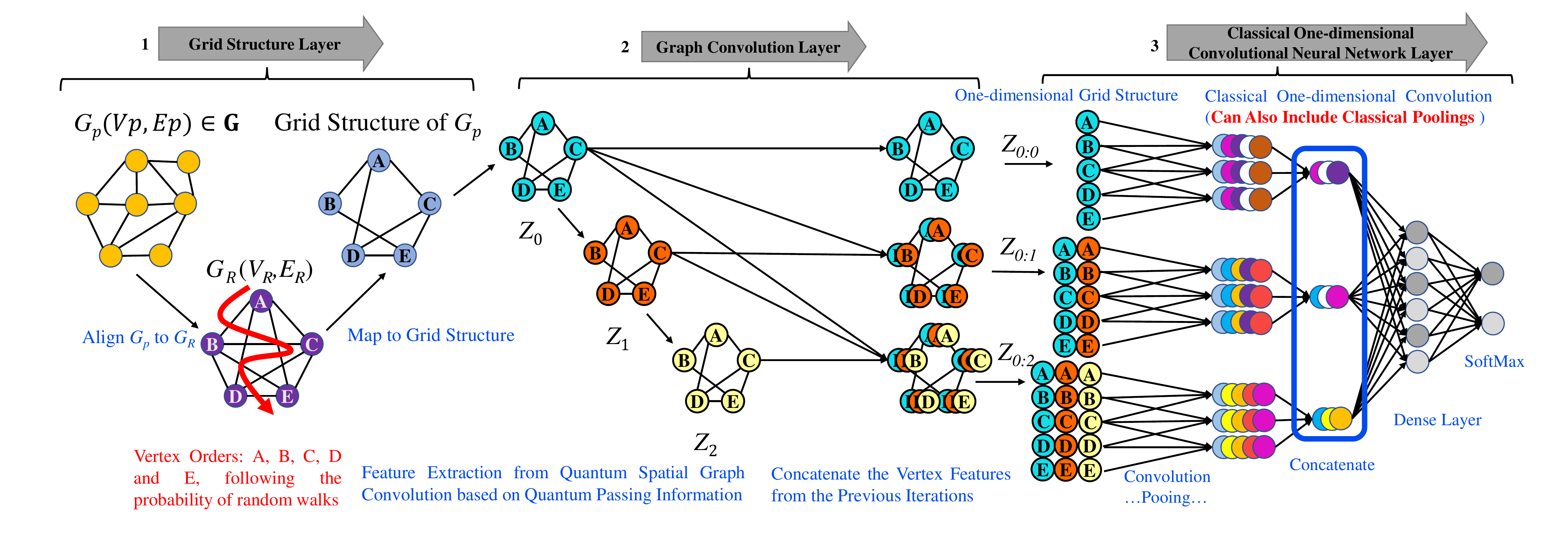}
 \vspace{-20pt}
  \caption{The architecture of the proposed QSGCNN model. An input graph $G_p(V_p,E_p)\in \mathbf{G}$ of arbitrary size is first aligned to the prototype graph $G_R(V_R,E_R)$. Then, $G_p$ is mapped into a fixed-sized aligned vertex grid structure, where the vertex orders follows that of $G_R$. The grid structure of $G_p$ is passed through multiple quantum spatial graph convolution layers to extract multi-scale vertex features, where the vertex information is propagated between specified vertices following the average mixing matrix. Since the graph convolution layers preserve the original vertex orders of the input grid structure, the concatenated vertex features through the graph convolution layers form a new vertex grid structure for $G_p$. This vertex grid structure is then passed to a traditional CNN layer to learn a classification function. Note, vertex features are visualized as different colors.}\label{f:QSGCNN}
 \vspace{-20pt}
\end{figure*}

In this subsection, we briefly review the concept of continuous-time quantum walks. Specifically, we use the average mixing matrix to capture the time-averaged behaviour of the quantum walk and to measure the quantum information being transmitted between the graph vertices. The continuous-time quantum walk is the quantum analogue of the continuous-time classical random walk~\cite{farhi1998quantum}, where the latter models a Markovian diffusion process over the vertices of a graph through the transitions between adjacent vertices. Let a sample graph be denoted as $G(V,E)$ with vertex set $V$ and edge set $E$. Like the classical random walk, the state space of the quantum walk is the vertex set $V$. Its state at time $t$ is a complex linear combination of the basis states $\Ket{u}$, i.e., $\Ket{\psi(t)} = \sum_{u\in V} \alpha_u(t) \Ket{u}$, where $\alpha_u (t) \in \mathbb{C}$ and $\Ket{\psi(t)} \in \mathbb{C}^{|V|}$ are the amplitude and both complex. Furthermore, $\alpha_u (t) \alpha_u^* (t)$ indicates the probability of the walker visiting vertex $u$ at time $t$, $\sum_{u \in V} \alpha_u (t) \alpha^{*}_u(t) = 1$, and $\alpha_u (t) \alpha^{*}_u(t) \in [0,1]$, for all $u \in V$, $t \in \mathbb{R}^{+}$. Unlike the classical counterpart, the continuous-time quantum walk evolves based on the Schr\"{o}dinger equation
\begin{equation}
\partial/\partial t \Ket{\psi_t} = -iH\Ket{\psi_t},
\end{equation}
where $H$ represents the system Hamiltonian. In this work, we use the adjacency matrix as the Hamiltonian. The behaviour of a quantum walk over the graph $G(V,E)$ at time $t$ can be summarized using the mixing matrix~\cite{godsil2013average}
\begin{align}\label{}
Q_M(t) = U(t) \circ U(-t) = e^{iHt} \circ e^{-iHt},
\end{align}
where the operation symbol $\circ$ represents the Schur-Hadamard product of $e^{iHt}$ and $e^{-iHt}$. Because $U$ is unitary, $Q_M(t)$ is a doubly stochastic matrix and each entry $Q_M(t)_{uv}$ indicates the probability of the walk visiting vertex $v$ at time $t$ when the walk initially starts from vertex $u$. However, $M(t)$ cannot converge, because $U(t)$ is also norm-preserving. To overcome this problem, we can enforce convergence by taking a time average. Specifically, we take the Ces\`{a}ro mean and define the average mixing matrix as $Q = \lim_{T \rightarrow \infty} \int_{0}^{T} Q_M(t) dt$,
where each entry $Q_{v_iv_j}$ of the average mixing matrix $Q$ represents the average probability for a quantum walk to visit vertex $v_j$ starting from vertex $v_i$, and $Q$ is still a doubly stochastic matrix. Furthermore, Godsil~\cite{godsil2013average} has indicated that the entries of $Q$ are rational numbers. We can easily compute $Q$ from the spectrum of the Hamiltonian. Specifically, let the adjacency matrix $A$ of $G$ be the Hamiltonian $H$. Let $\lambda_1,\ldots,\lambda_{|V|}$ represent the $|V|$ distinct eigenvalues of $H$ and $\mathbb{P}_j$ is the matrix representation of the orthogonal projection on the eigenspace associated with the $\lambda_j$, i.e., $H = \sum_{j = 1}^{|V|} \lambda_j \mathbb{P}_j.$ Then, we can re-write the average mixing matrix $Q$ as
\begin{equation}
Q = \sum_{j = 1}^{|V|} \mathbb{P}_j \circ \mathbb{P}_j.
\end{equation}

\subsection{Transitive Alignment Between Vertices of Graphs}
We introduce a new transitive vertex alignment method. To this end, we commence by identifying a family of prototype representations that reflect the main characteristics of the vectorial vertex representations over a set of graphs $\mathbf{G}$. Assume there are $n$ vertices over all graphs in $\mathbf{G}$, and the associated $K$-dimensional vectorial representations of these vertices are $\mathbf{{R}}^K =(\mathrm{R}_1^K,\mathrm{R}_2^K,\ldots,\mathrm{R}_n^K)$. We use $k$-means~\cite{witten2011data} to identify $M$ centroids over all representations in $\mathbf{{R}}^K$. Specifically, given $M$ clusters $\Omega=(c_1,c_2,\ldots,c_M)$, the aim of $k$-means is to minimize the objective function
\begin{equation}
\arg\min_{\Omega}  \sum_{i=1}^M \sum_{\mathrm{R}_j^K \in c_i^K} \|\mathrm{R}_j^K- \mu_i^K\|^2,\label{kmeans}
\end{equation}
where $\mu_i^K$ is the mean of the vectorial vertex representations belonging to the $i$-th cluster $c_i$. Since Eq.(\ref{kmeans}) minimizes the sum of the square Euclidean distances between the vertex points $\mathrm{R}_j^K$ and the centroid point of cluster $c_i^K$, the $M$ centroid points $\{\mu_1^K,\cdots,\mu_i^K,\cdots,\mu_M^K\}$ can be seen as a family of $K$-dimensional \textbf{prototype representations} that encapsulate representative characteristics over all graphs in $\mathbf{G}$.

Let $\mathbf{G}=\{G_1,\cdots,G_p,\cdots,G_q,\cdots,G_N\}$ be a set of graphs. For each graph $G_p(V_p,E_p)\in {\mathbf{G}}$ and each vertex $v_i\in V_p$ associated with its $K$-dimensional vectorial representation $\mathrm{{R}}_{p;i}^K$, we commence by identifying the set of $K$-dimensional prototype representations as $\mathbf{PR}^K=\{\mu_1^K,\ldots,\mu_j^K,\ldots,\mu_M^K \}$ for the graph set $\mathbf{G}$. To establish a set of correspondences between the graph vertices, we align the vectorial vertex representations of each graph $G_p$ to the family of prototype representations $\mathbf{PR}^K$. The alignment process is similar to that introduced in~\cite{DBLP:conf/icml/Bai0ZH15} for point matching in a pattern space. Specifically, we compute a $K$-level affinity matrix in terms of the Euclidean distances between the two sets of points
\begin{align}
A^K_p(i,j)=\|\mathrm{{DB}}_{p;i}^K - \mu_j^K\|_2.\label{AffinityM}
\end{align}
where $A^K_p$ is a ${|V_p|}\times {M}$ matrix, and each element $R^K_p(i,j)$ represents the distance between the vectrial representation $\mathrm{{R}}_{p;i}^K$ of $v_\in V_p$ and the $j$-prototype representation $\mu_j^K\in \mathbf{PR}^K$. If the value of $A^K_p(i,j)$ is the smallest in row $i$, we say that $\mathrm{{R}}_{p;i}^K$ is aligned to $\mu_j^K$, i.e., the vertex $v_i$ is aligned to the $j$-th prototype representation. Note that for each graph there may be two or more vertices aligned to the same prototype representation. We record the correspondence information using the $K$-level correspondence matrix
$C^K_p\in \{0,1\}^{|V_p|\times M}$
\begin{equation}
C^K_p(i,j)=\left\{
\begin{array}{cl}
1   & \small{\mathrm{if} \  A^K_p(i,j) \ \mathrm{is \ the \ smallest \ in \ row } \ i} \\
0   & \small{\mathrm{otherwise}}.
\end{array} \right.
\label{CoMatrix}
\end{equation}

For a pair of graphs $G_p$ and $G_q$, if their vertices $v_p$ and $v_q$ are aligned to the same prototype representation $\mathrm{PR}_j^K$, we say that $v_p$ and $v_q$ are also aligned. Thus, we can identify the transitive alignment information between the vertices of all graphs in $\mathbf{G}$, by matching their vertices to a common set of reference points, i.e., the prototype representations.

\noindent\textbf{Discussion:} The alignment process illustrated by Eq.(\ref{AffinityM}) and Eq.(\ref{CoMatrix}) can be explained by the objective function of $k$-means defined by Eq.(\ref{kmeans}). This is because identifying the smallest element $A^K_p(i,j)$ in the $i$-row of $A^K_p$ is equivalent to assigning the vectorial representation $\mathrm{{R}}_{p;i}^K$ of $v_i\in V_p$ to the cluster $c_i^K$ whose mean vector is $\mu_i^K$. As a result, the proposed alignment procedure can be seen as an optimization process that gradually minimizes the inner-vertex-cluster sum of squares over the vertices of all graphS, and can establish reliable vertex correspondence information over all graphs.

\section{Quantum Spatial Graph Convolutional Neural Network}\label{s3}

In this section, we develop a new Quantum Spatial Graph Convolutional Neural Network (QSGCNN) model. The architecture of the proposed model is shown in Fig.\ref{f:QSGCNN}. Specifically, the architecture is composed of three sequential stages, i.e., 1) the grid structure construction and input layer, 2) the quantum spatial graph convolution layer, and 3) the traditional convolutional neural network and Softmax layers. Specifically, the grid structure construction and input layer a) first maps graphs of arbitrary sizes into fixed-sized grid structures with consistent vertex orders, and b) inputs the grid structures into the proposed QSGCNN model. With the input graph grid structures to hand, the quantum spatial graph convolution layer further extracts multi-scale vertex features by propagating vertex feature information between the aligned grid vertices. Since the extracted vertex features from the graph convolution layer preserve the original vertex orders of the input grid structures, the traditional convolutional neural network and Softmax layer can read the extracted vertex features and predict the graph class. 

\subsection{Aligned Vertex Grid Structures of Graphs}
In this subsection, we show how to map graphs of different sizes onto fixed-sized aligned vertex grid structures and associated corresponding fixed-sized aligned grid vertex adjacency matrices. For the set of graphs $\mathbf{G}$ defined earlier, suppose $G_p(V_p,E_p,A_p)\in \mathbf{G}$ is a sample graph, with $V_p$ representing the vertex set, $E_p$ representing the edge set, and $A_p$ representing the vertex adjacency matrix. Suppose each vertex $v_p\in V_p$ is represented as a $c$-dimensional feature vector. Then the features of all the $n$ vertices can be encoded using the $n\times c$ matrix $X_p$, i.e., $X_p\in \mathbb{R}^{n\times c}$. Note that the row of $X_p$ follows the same vertex order of $A_p$. If the graphs in $\mathbf{G}$ are vertex attributed graphs, $X_p$ can be the one-hot encoding matrix of the vertex labels. For unattributed graphs, we propose to use the vertex degree as the vertex label. Based on the transitive vertex alignment method introduced in Section~\ref{s2}, for each graph $G_p\in \mathbf{G}$, we commence by computing the $K$-level vertex correspondence matrix $C^K_p$ that records the correspondence information between the $K$-dimensional vectorial vertex representation of $G_p$ and the $K$-dimensional prototype representations in $\mathbf{PR}^K=\{\mu_1^K,\ldots,\mu_j^K,\ldots,\mu_M^K \}$ of $\mathbf{G}$. The row and column of $C^K_p$ are indexed by the vertices in $V_p$ and the prototype representations in $\mathbf{PR}^K$, respectively. With $C^K_p$ to hand, we compute the $K$-level aligned vertex feature matrix for $G_p$ as
\begin{equation}
\widehat{X}_{p}^{K}= (C^K_p)^T X_p,\label{alignDB}
\end{equation}
where $\widehat{X}_{p}^{K}$ is a $M\times c$ matrix and each row of $\widehat{X}_{p}^{K}$ represents the feature of a corresponding aligned vertex. Moreover, we also compute the associated $K$-level aligned vertex adjacency matrix for $G_p$ as
\begin{equation}
\widehat{A}_{p}^{K}= (C^K_p)^T (A_{p}) (C^K_p),\label{alignA}
\end{equation}
where $\widehat{A}_{p}^{K}$ is a $M\times M$ matrix. With the correspondence matrix $C^K_p$ to hand, $\widehat{X}_{p}^{K}$ and $\widehat{A}_{p}^{K}$ are computed from the original vertex feature matrix and adjacency matrix, respectively, by mapping the original feature and adjacency information of each vertex $v_p\in V_p$ to that of the new aligned vertices indexed by the corresponding prototypes in $\mathbf{PR}^K$. In other words $\widehat{X}_{p}^{K}$ and $\widehat{A}_{p}^{K}$ encapsulate the original feature and structural information of $G_p$. Note also that according to Eq.~\ref{CoMatrix} each vertex $v_p\in V_p$ can be aligned to more than one prototype, and thus in general $\widehat{A}_{p}^{K}$ is a weighted adjacency matrix.

In order to construct the fixed-sized aligned grid structure for each graph $G_p\in \mathbf{G}$, we need to establish a consistent order for the vertices of the graphs in $\mathbf{G}$. Since the vertices of all the graphs are aligned to the same prototype representations, we determine the vertex orders by reordering the prototype representations. To this end, we construct a prototype graph that captures the pairwise similarity between the prototype representations. Given this graph, one approach could be to sort the prototype representations based on their degree. This would be equivalent to sorting the prototypes in orders of average similarity to the remaining ones. Specifically, we compute the prototype graph $G_{\mathrm{R}}(V_{\mathrm{R}},E_{\mathrm{R}})$ that characterizes the relationship information between the $K$-dimensional prototype representations in $\mathbf{PR}^K$, with each vertex $v_j\in V_{\mathrm{R}}$ representing the prototype representation $\mu_j^K\in \mathbf{PR}^K$ and each edge $(v_j,v_k)\in E_{\mathrm{R}} $ representing the similarity between $\mu_j^K\in \mathbf{PR}^K$ and $\mu_k^K\in \mathbf{PR}^K$. The similarity between two vertices of $G_{\mathrm{R}}$ is computed as \begin{equation}
s(\mu_j^K,\mu_k^K)=\exp (-\frac{\| \mu_j^K-\mu_k^K   \|_2}{K}).
\end{equation}
The degree of each prototype representation $\mu_j^K$ is $D_R(\mu_j^K)=\sum_{k=1}^{M}s(\mu_j^K,\mu_k^K)$. We sort the $K$-dimensional prototype representations in $\mathbf{PR}^K$ according to their degree $D_R(\mu_j^K)$. Then, we rearrange $\widehat{X}_{p}^{K}$ and $\widehat{A}_{p}^{K}$ accordingly.
\begin{figure*}
 \vspace{-0pt}
 \centering
\includegraphics[width=0.85\linewidth]{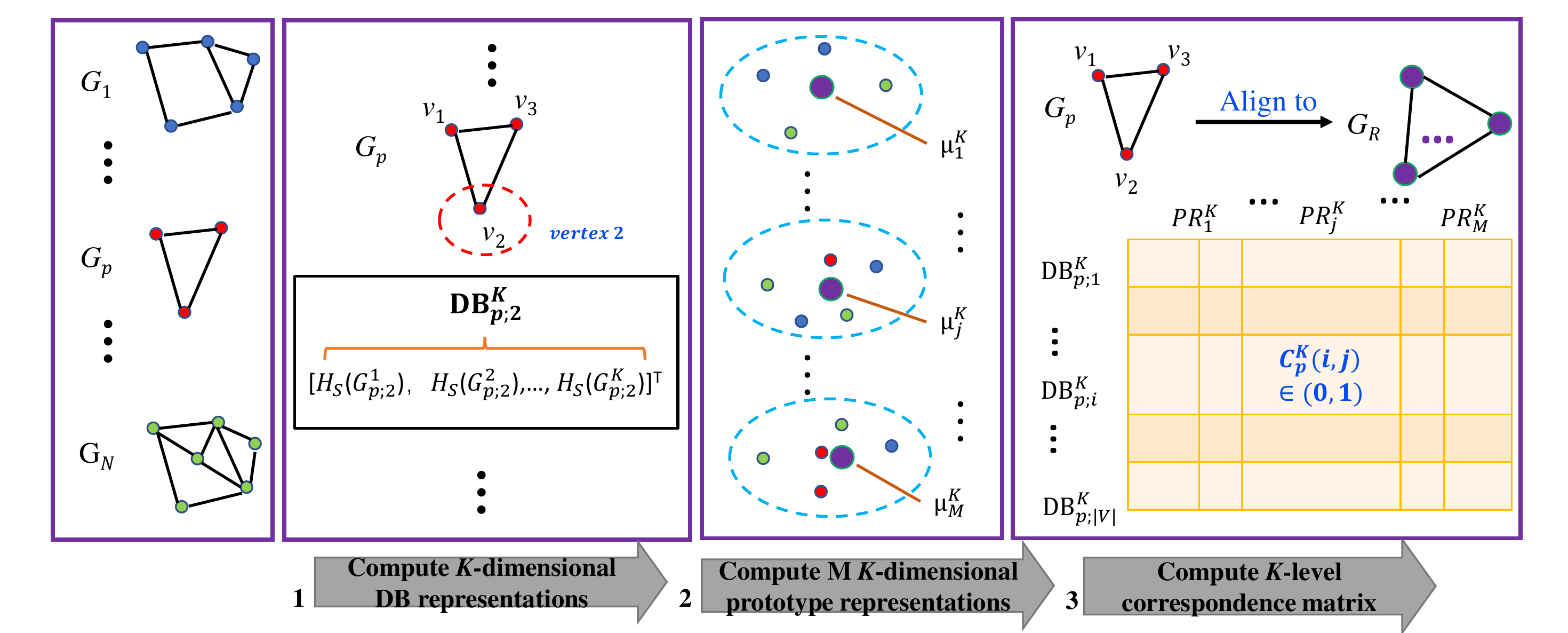}
 \vspace{-10pt}
  \caption{The procedure of computing the correspondence matrix. Given a set of graphs, for each graph $G_p$: (1) we compute the $K$-dimensional depth-based (DB) representation $\mathrm{{DB}}_{p;v}^K$ rooted at each vertex (e.g., vertex 2) as the $K$-dimensional vectorial vertex representation, where each element $H_s(G_{p;2}^K)$ represents the Shannon entropy of the $K$-layer expansion subgraph rooted at vertex $v_2$ of $G_p$ associated with steady state random walk~\cite{DBLP:journals/prl/Bai18}; (2) we identify a family of $K$-dimensional prototype representations $\mathbf{PR}^K = \{\mu_1^K,\ldots,\mu_j^K,\ldots,\mu_M^K \}$ using k-means on the $K$-dimensional DB representations of all graphs; (3) we align the $K$-dimensional DB representations to the $K$-dimensional prototype representations and compute a $K$-level correspondence matrix $C_p^K$.}\label{f:alignment}
 \vspace{-10pt}
\end{figure*}

Finally, note that, to construct reliable grid structures for graphs, in this work we employ the depth-based representations as the vectorial vertex representations to compute the required $K$-level vertex correspondence matrix $C_p^K$. Specifically, the depth-based representation of each vertex is computed by measuring the entropies on a family of $k$-layer expansion subgraphs rooted at the vertex~\cite{DBLP:journals/prl/Bai18}, where the parameter $k$ varies from $1$ to $K$. Moreover, it has been shown that such a $K$-dimensional depth-based representation of a vertex can be seen as \emph{\textbf{a nested vertex representation}} that encapsulates rich nested entropy-based information content flow from each local vertex to the global graph structure~\cite{DBLP:journals/prl/Bai18}, as a function of depth. The process of computing the correspondence matrix $C_p^K$ associated with depth-based representations is
shown in Fig.\ref{f:alignment}. When we vary the largest layer $K$ of the expansion subgraphs from $1$ to $L$ (i.e., $K\leq L$), we compute the final \textbf{aligned vertex grid structure} for each graph $G_p\in \mathbf{G}$ as
\begin{equation}
\widehat{X}_{p}= \sum_{K=1}^L \frac{\widehat{X}_{p}^{K}}{L},\label{AlignV}
\end{equation}
and the associated \textbf{aligned grid vertex adjacency matrix} as
\begin{equation}
\widehat{A}_{p}= \sum_{K=1}^L \frac{\widehat{A}_{p}^{K}}{L},\label{AlignA}
\end{equation}
where $\widehat{X}_{p}$ is a $M\times c$ matrix, and $\widehat{A}_{p}$ is a $M\times M$ matrix.
\\
\noindent\textbf{Discussion:} Eq.(\ref{AlignV}) and Eq.(\ref{AlignA}) transform the original graphs $G_p\in \mathbf{G}$ with varying number of nodes $|V_p|$ into a new aligned grid graph structure with the same number of vertices, where $\widehat{X}_{p}$ is the corresponding aligned grid vertex feature matrix and $\widehat{A}_{p}$ is the corresponding aligned grid vertex adjacency matrix. Since for any graph $G_p\in \mathbf{G}$ the rows of $\widehat{X}_{p}$ are consistently indexed by the same prototype representations, the fixed-sized vertex grid structure $\widehat{X}_{p}$ can be directly employed as the input of a traditional convolutional neural network. In other words, one can apply a fixed sized classical convolutional filter to slide over the rows of $\widehat{X}_{p}$ and learn the feature for $G_p\in \mathbf{G}$. Finally, note that $\widehat{X}_{p}$ and $\widehat{A}_{p}$ accurately encapsulate the original feature and structural information of $G_p$, respectively.

\subsection{The Quantum Spatial Graph Convolution Layer}

In this subsection, we propose a new quantum spatial graph convolution layer to further extract the features of the vertices of each graph. This is defined by quantum information propagation between aligned grid vertices. To this end, we employ the average mixing matrix of the continuous-time quantum walk on the associated aligned grid vertex adjacency matrix. For the sample graph $G_p(V_p,E_p)$, we pass the aligned vertex grid structure $\widehat{X}_{p}\in \mathbb{R}^{M\times c}$ and the associated aligned grid vertex adjacency matrix $\widehat{A}_{p}\in \mathbb{R}^{M\times M}$ of $G_p$ as the input of the quantum spatial graph convolution layer. The proposed spatial graph convolution layer takes the following form, i.e.,
\begin{equation}
Z= \mathrm{Relu}(Q \widehat{X}_{p} W),\label{GCN_EQ}
\end{equation}
where $\mathrm{Relu}$ is the rectified linear units function (i.e., a nonlinear activation function), $Q$ is the average mixing matrix of the continuous-time quantum walk on $\widehat{A}_{p}$ of $G_p$ defined in Section~\ref{s2.1}, $W\in \mathbb{R}^{c\times c^{'}}$ is the matrix of trainable parameters of the proposed graph convolutional layer, and $Z\in \mathbb{R}^{M\times c^{'}}$ is the output activation matrix.

The proposed quantum spatial graph convolution layer defined by Eq.(\ref{GCN_EQ}) consists of three steps. In the first step the operation $\widehat{X}_{p} W$ is applied to transform the aligned grid vertex information matrix into a new aligned grid vertex information matrix. This in turn maps the $c$-dimensional features of each aligned grid vertex into new $c^{'}$-dimensional features, i.e., $\widehat{X}_{p} W$ maps the $c$ feature channels to $c^{'}$ channels in the next layer. The weights $W$ are shared among all aligned grid vertices. The second step computes $Q Y$, where $Y:= \widehat{X}_{p} W$. This propagates the feature information of each aligned grid vertex to the remaining vertices as well as the vertex itself, in terms of the vertex visiting information of quantum walks. Specifically, we note that ${Q}_{ij}$ encapsulates the average probability for a continuous-time quantum walk to visit the $j$-th aligned grid vertex starting from the $i$-th aligned grid vertex, and $(Q \widehat{X}_{p}^{'})_i=\sum_j {Q}_{ij} {Y}_{j}$. Here, $i$ can be equal to $j$, i.e., $Q$ includes the self-loop information for each vertex. Thus, the $i$-th row of the resulting matrix of $Q \widehat{X}_{p}^{'}$ is the feature summation of the $i$-th aligned grid vertex and the remaining aligned grid vertices associated with the average visiting probability of quantum walks from the $i$-th vertex to the remaining vertices as well as the $i$-th vertex itself. The final step applies the rectified linear unit function to $Q \widehat{X}_{p} W)$ and outputs the graph convolution result.

The proposed quantum spatial graph convolution propagates the aligned grid vertex information in terms of the vertex visiting information associated with the continuous-time quantum walk between vertices. To further extract the multi-scale features of the aligned grid vertices, we stack multiple graph convolution layers defined by Eq.(\ref{GCN_EQ}) as follows
\begin{equation}
Z_{t+1}= \mathrm{Relu}(Q Z_t W_t),\label{GCN_EQM}
\end{equation}
where $Z_0$ is the input aligned vertex grid structure $\widehat{X}_{p}$, $Z_t \in \mathbb{R}^{M\times c_t}$ is the output of the $t$-th spatial graph convolution layer, and $W_t \in \mathbb{R}^{c_t\times c_{t+1}}$ is the trainable parameter matrix mapping $c_t$ channels to $c_{t+1}$ channels.

After each $t$-th quantum spatial graph convolutional layer, we also add a layer to horizontally concatenate the output $Z^t$ associated with the outputs of the previous $1$ to $t-1$ spatial graph convolutional layers as well as the original input $Z^0$ as $Z_{0:t}$, i.e., $Z_{0:t}=[Z_0,Z_1,\ldots,Z^t]$ and $Z_{0:t}\in  \mathbb{R}^{M\times \sum_{z=0}^t c_z}$. As a result, for the concatenated output $Z_{0:t}$, each of its row can be seen as the new multi-scale features for the corresponding grid vertex.\\

\noindent\textbf{Discussion:} Note that the proposed quantum spatial graph convolution only extracts new features for the grid vertex and does not change the orders of the vertices. As a result, both the output $Z^t$ and the concatenated output $Z_{0:t}$ preserve the grid structure property of the original input $Z_0=\widehat{X}_{p}$, and can be directly employed as the input of the traditional convolutional neural network. This provides an elegant way of bridging the gap between the proposed quantum spatial graph convolution layer and the traditional convolutional neural network, making an end-to-end deep learning architecture that integrates the graph representation and
learning in both the quantum spatial graph convolution layer and the traditional convolution layer for graph classification problems.

\subsection{The Traditional Convolutional Neural Network Layers}
After the $t$-th proposed quantum spatial graph convolution layers, we get a concatenated vertex grid structure $Z_{0:t}\in  \mathbb{R}^{M\times \sum_{z=0}^t c_z}$, where each row of $Z_{0:t}$ represents the multi-scale feature for a corresponding grid vertex. As we mentioned above, each grid structure $Z_{0:t}$ can be directly employed as the input to the traditional convolutional neural network (CNN). Specifically, the Classical One-dimensional CNN part of Fig.\ref{f:QSGCNN} exhibits the architecture of the traditional CNN layers associated with each $Z_{0:t}$. Here, each concatenated vertex grid structure $Z_{0:t}$ is seen as a $M \times 1$ (in Fig.\ref{f:QSGCNN} $M = 5$) vertex grid structure and each vertex is represented by a $\sum_{z=0}^t c_z$-dimensional feature, i.e., the channel of each grid vertex is $\sum_{z=0}^t c_z$. Then, we add a one-dimensional convolutional layer. The convolutional operation can be performed by sliding a fixed-sized filter of size $k\times 1$ (in Fig.\ref{f:QSGCNN} $k = 3$) over the spatially neighboring vertices. After this, several MaxPooling layers and remaining one-dimensional convolutional layers can be added to learn the local patterns on the aligned grid vertex sequence. Finally, when we vary $t$ from $0$ to $T$ (in Fig.\ref{f:QSGCNN} $T = 2$), we will obtain $T+1$ extracted pattern representations. We concatenate the extracted patterns of each $Z_{0:t}$ and add a fully-connected layer followed by a Softmax layer.

\subsection{Discussion of the Proposed QSGCNN Model}\label{s3.4}

The proposed QSGCNN model is related to some existing state-of-the-art graph convolution network models and graph kernels. However, there are a number of significant theoretical differences between the proposed QSGCNN model and these state-of-the-art methods, explaining the effectiveness of the proposed model. In this subsection, we discuss the relationships between these methods and demonstrate the advantages of the proposed model.

\textbf{First}, similar to the quantum spatial graph convolution of the proposed QSGCNN model, the associated graph convolution of the Deep Graph Convolutional Neural Network (DGCNN)~\cite{DBLP:conf/aaai/ZhangCNC18} and the spectral graph convolution of the Fast Approximate Graph Convolutional Neural Network (FAGCNN)~\cite{DBLP:journals/corr/KipfW16} also propagate the features between the graph vertices. Specifically, the graph convolutions of the DGCNN and FAGCNN models use the graph adjacency matrix or the normalized Laplacian matrix to determine how to pass the information among the vertices. In contrast, our quantum spatial graph convolution utilizes the average mixing matrix of the continuous-time quantum walk associated with the graph. As we mentioned in Section~\ref{s2.1}, the quantum walk is not dominated by the low frequency values of the Laplacian spectrum and thus has a better ability to distinguish different graph structures. As a result, the proposed method can extract more discriminative vertex features.

\textbf{Second}, in order to maintain the scale of the vertex features after each graph convolution layer, the graph convolution of the DGCNN model~\cite{DBLP:conf/aaai/ZhangCNC18} and the spectral graph convolution of the FAGCNN model~\cite{DBLP:journals/corr/KipfW16} need to perform a multiplication by the inverse of the vertex degree matrix. For instance, the graph convolution layer of the DGCNN model associated with a graph having $n$ vertices is
\begin{equation}
Z= \mathrm{f}(\widetilde{D}^{-1} \widetilde{A} X W),\label{GCN_EAM}
\end{equation}
where $\widetilde{A}=A+I$ is the adjacency matrix of the graph with added self-loops, $\widetilde{D}$ is the degree matrix of $\widetilde{A}$, $X^{n\times c}$ is the vertex feature matrix with each row representing the $c$-dimensional features of a vertex, $W^{c\times c^{'}}$ is the matrix of trainable parameters, $\mathrm{f}$ is a nonlinear activation function (e.g., the Relu function), and $Z^{n\times c^{'}}$ is the output. In a manner similar to the proposed quantum spatial graph convolution defined in Eq.(\ref{GCN_EQ}), $X W$ maps the $c$-dimensional features of each vertex into a set of new $c^{'}$-dimensional features. Moreover, $\widetilde{A} Y$ ($Y:= \widehat{X}_{p} W$) propagates the feature information of each vertex to its neighboring vertices as well as the vertex itself. The $i$-th row $(\widetilde{A} Y)_i$ of the resulting matrix $\widetilde{A} Y$ represents the extracted features of the $i$-th vertex, and corresponds to the summation of $Y_i$ itself and $Y_j$ from the neighbor vertices of the $i$-th vertex. Multiplying by the inverse of $\widetilde{D}$ (i.e., $\widetilde{D}^{-1}$) can be seen as the process of normalizing and assigning equal weights between the $i$-th vertex and each of its neighbours. In other words, the graph convolution of the DGCNN model considers the mutual-influences between specified vertices for the convolution operation as the same. In contrast, the quantum spatial graph convolution of the proposed QSGCNN model defined in Eq.(\ref{GCN_EQ}) assigns an average quantum walk visiting probability distribution to specified vertices with each vertex having a different visiting probability as the weight. Therefore, the extracted vertex feature is the weighted summation of the specified vertex features. As a result, the quantum spatial graph convolution of the proposed QSGCNN model not only maintains the feature scale, but also discriminates the mutual-influences between specified vertices in terms of the different visiting probabilities during the convolution operation.

\textbf{Third}, similar to the proposed QSGCNN model, both the PATCHY-SAN based Graph Convolution Neural Network (PSGCNN) model~\cite{DBLP:conf/icml/NiepertAK16} and the DGCNN model~\cite{DBLP:conf/aaai/ZhangCNC18} need to rearrange the vertex order of each graph structure and transform each graph into the fixed-sized vertex grid structure. Specifically, the PSGCNN model first forms the grid structures and then performs the standard classical CNN on the grid structures. The DGCNN model sorts the vertices through a SortPooling associated with the extracted vertex features from multiple spatial graph convolution layers. Unfortunately, both the PSGCNN model and the DGCNN model sort the vertices of each graph based on the local structural descriptor, ignoring consistent vertex correspondence information between different graphs. By contrast, the proposed QSGCNN model associates with a transitive vertex alignment procedure to transform each graph into an aligned fixed-sized vertex grid structure. As a result, only the proposed QSGCNN model can integrate the precise structural correspondence information over all graphs under investigations.

\textbf{Fourth}, when the PSGCNN model~\cite{DBLP:conf/icml/NiepertAK16} and the DGCNN model~\cite{DBLP:conf/aaai/ZhangCNC18} form fixed-sized vertex grid structures, some vertices with lower ranking will be discarded. Moreover, the Neural Graph Fingerprint Network (NGFN)~\cite{DBLP:conf/nips/DuvenaudMABHAA15} and the Diffusion Convolution Neural Network (DCNN)~\cite{DBLP:conf/nips/AtwoodT16} tend to capture global-level graph features by summing up the extracted local-level vertex features through a SumPooling layer, since both the NGFN model and the DCNN model cannot directly form vertex grid structures. This leads to significant information loss for local-level vertex features. By contrast, the required aligned vertex grid structures and the associated grid vertex adjacency matrices for the proposed QSGCNN model can accurately encapsulate both the original vertex features and the topological structure information of the original graphs. As a result, the proposed QSGCNN overcomes the shortcoming of information loss arising in the mentioned state-of-the-art graph convolutional neural network models.


\textbf{Fifth}, similar to the DGCNN model~\cite{DBLP:conf/aaai/ZhangCNC18}, the quantum spatial graph convolution of the proposed QSGCNN model is also related to the Weisfeiler-Lehman subtree kernel (WLSK)~\cite{shervashidze2010weisfeiler} Specifically, the WLSK kernel employs the classical Weisfeiler-Lehman (WL) algorithm as a canonical labeling method to extract multi-scale vertex features corresponding to subtrees for graph classification. The key idea of the WL method is to concatenate a vertex label with the labels of its neighbor vertices, and then sort the concatenated label lexicographically to assign each vertex a new label. The procedure repeats until a maximum iteration $h$, and each vertex label at an iteration $h$ corresponds to a subtree of height $t$ rooted at the vertex. If the concatenated label of two vertices are the same, the subtree rooted at the two vertices are isomorphic, i.e., the two vertices are seen to share the same structural characteristics within the graph. The WLSK kernel uses this idea to measure the similarity between two graphs. It uses the WL method to update the vertex labels, and then counts the number of identical vertex labels (i.e. counting the number of the isomorphic subtrees) until the maximum of the iteration $h$ in order to compare two graphs at multiple scales. To exhibit the relationship between the proposed quantum spatial graph convolution defined in Eq.(\ref{GCN_EQ}) and the WLSK kernel, we decompose Eq.(\ref{GCN_EQ}) in a row-wise manner, i.e.,
\begin{equation}
Z_i= \mathrm{Relu}(Q_i Y)= \mathrm{Relu} (Q_{ii}Y_i+ \sum_j Q_{ij}),\label{GCN_WL}
\end{equation}
where $Y=\widehat{X}_{p} W$. For Eq.(\ref{GCN_WL}), $Y_i$ can be seen as the continuous valued vectorial vertex label of the $i$-th vertex. Moreover, if $Q_{ij}>0$, the quantum walk starting from the $i$-th vertex can visit the $j$-th vertex, and the visiting probability is $Q_{ij}$. In a manner similar to the WL methods, Eq.(\ref{GCN_WL}) aggregates the continuous label $Y_i$ of the $i$-th vertex and the continuous labels $Y_j$ of the vertices, that can be visited by the quantum walk starting from the $i$-th vertex, as a new signature vector $Q_{ii}Y_i+ \sum_j Q_{ij}$ for the $i$-th vertex. The $\mathrm{Relu}$ function maps $Q_{ii}Y_i+ \sum_j Q_{ij}$ to a new continuous vectorial label. As a result, the the quantum spatial graph convolution of the proposed QSGCNN model can be seen as \textbf{a quantum version of the WL algorithm}, in terms of the quantum vertex information propagation formulated by the quantum walk. As we mentioned in Section~\ref{s2.1}, the quantum walk can significantly reduce the effect of the tottering problem. On the other hand, the classical WL method also suffers from tottering problem~\cite{DBLP:conf/icml/Bai0ZH15}. As a result, the quantum spatial graph convolution can address the tottering problem arising in the classical WL method, and the graph convolution of the DGCNN model is similar to the clasical WL method. In other words, the quantum spatial graph convolution of the proposed QSGCNN model can learn better vertex features of graphs

\textbf{Finally}, note that the proposed QSGCNN model for each graph is invariant with respect to the permutation of the vertices, indicating that the activations of a pair of isomorphic graphs will be the same. As we mentioned, the proposed QSGCNN model consists of three stages, i.e., a) the grid structure construction and input layer, b) the quantum spatial graph convolution layer, and c) the traditional CNN layer. For the first layer, the construction of grid structures relies on the vertex features and adjacency matrix, and is invariant to vertex permutations. As a result, the grid structures for a pair of isomorphic graphs are the same. For the second layer, the input grid structures of different graphs share the same parameter weights, thus the quantum spatial graph convolutions will produce the same extracted vertex features for a pair of isomorphic graphs associated with the same grid structures. Consequently, the subsequent classical CNN layer will correctly identify the isomorphic graphs. As a result, the proposed QSGCNN model can correctly identify pairs of isomorphic graphs.

These observations reveal the advantages of the proposed QSGCNN model, explaining the effectiveness of the proposed model. The proposed QSGCNN model not only overcomes the shortcomings of existing state-of-the-art methods, but also bridges the theoretical and computational gaps between these methods.

\begin{table*}
\centering {
\scriptsize
\vspace{-0pt}
\caption{Information of the Graph Datasets}\label{T:GraphInformation} \vspace{0pt}
\begin{tabular}{|c||c||c||c||c||c||c||c||c||c|}

  \hline
 ~Datasets ~          & ~MUTAG  ~ & ~NCI1~     & ~PROTEINS~& ~D\&D~       & ~PTC(MR)~  & ~COLLAB  ~ & ~IMDB-B~     & ~IMDB-M~  & ~RED-B~\\ \hline \hline

  ~Max \# vertices~   & ~$28$~    & ~$111$~    & ~$620$~   &  ~$5748$~    & ~$109$~ & ~$492$~    & ~$136$~    & ~$89$~ & ~$3783$~\\ \hline


  ~Mean \# vertices~  & ~$17.93$~ & ~$29.87$~  & ~$39.06$~ &  ~$284.30$~  & ~$25.60$~ & ~$74.49$~ & ~$19.77$~  & ~$13.00$~ & ~$429.61$~\\  \hline

  ~Mean \# edges~     & ~$19.79$~ & ~$32.30$~  & ~$72.82$~ &  ~$715.65$~  & ~$14.69$~ & ~$4914.99$~ & ~$193.06$~  & ~$131.87$~ & ~$497.80$~\\  \hline

  ~\# graphs~         & ~$188$~   &  ~$4110$~  & ~$1113$~  &  ~$1178$~    & ~$344$~  & ~$5000$~   &  ~$1000$~  & ~$1500$~  & ~$2000$~    \\ \hline
~\# vertex labels~     & ~$7$~   &  ~$37$~  & ~$61$~  &  ~$82$~    & ~$19$~  & ~$-$~   &  ~$-$~  & ~$-$~   & ~$-$~   \\ \hline

~\# classes~        & ~$2$~     &  ~$2$~     & ~$2$~     &  ~$2$~       & ~$2$~    &  ~$3$~     & ~$2$~     &  ~$3$~   & ~$2$~   \\ \hline

~Description~     & ~Chemical~   &  ~Chemical~  & ~Chemical~  &  ~Chemical~    & ~Chemical~  & ~Social~   &  ~Social~  & ~Social~    & ~Social~   \\ \hline

\end{tabular}
} \vspace{-0pt}
\end{table*}

\begin{table*}
\centering {
\tiny
\caption{Classification Accuracy (In $\%$ $\pm$ Standard Error) for Comparisons with Graph Kernels.}\label{T:ClassificationGK}
\vspace{0pt}
\begin{tabular}{|c||c||c||c||c||c||c||c||c||c|}

  \hline
 ~Datasets~& ~MUTAG  ~        & ~NCI1~           & ~PROTEINS~         & ~D\&D~              & ~PTC(MR)~  & ~COLLAB~        & ~IBDM-B~        & ~IBDM-M~  & ~RED-B~\\ \hline \hline

 ~\textbf{QSGCNN}~  & ~$\textbf{90.52}\pm0.95$~& ~$77.50\pm0.91$~ & ~$\textbf{75.90}\pm0.79$~   &  ~$\textbf{81.70}\pm0.92$~   & ~$\textbf{63.37}\pm1.15$~ & ~$78.80\pm0.89$ ~& ~$\textbf{73.62}\pm1.12$ & ~$\textbf{51.60}\pm1.15$  & ~$\textbf{91.50}\pm0.24$\\ \hline

  ~JTQK~   & ~$85.50 \pm0.55$~& ~$\textbf{85.32}\pm0.14$~ & ~$72.86\pm0.41$~   &  ~$79.89\pm0.32$~   & ~$58.50\pm0.39$~  &~$76.85\pm0.40$~       &~$72.45\pm0.81$~          &  ~$50.33\pm0.49$~ & ~$77.60\pm0.35$\\ \hline

  ~WLSK~   & ~$82.88\pm0.57 $~ &~$84.77\pm0.13$~ & ~$73.52\pm0.43$~   &  ~$79.78\pm0.36$~   & ~$58.26\pm0.47 $~ &~$77.39\pm0.35$~       &~$71.88\pm0.77$~          &  ~$49.50\pm0.49$~ & ~$76.56\pm0.30$\\   \hline


  ~WL-OA~   & ~$82.88\pm0.57 $~ &~$84.77\pm0.13$~ & ~$73.52\pm0.43$~   &  ~$79.78\pm0.36$~   & ~$58.26\pm0.47 $~ &~$\textbf{80.70}\pm0.10$~       &~$71.88\pm0.77$~          &  ~$49.50\pm0.49$~ & ~$76.56\pm0.30$\\   \hline

  ~SPGK~   & ~$83.38\pm0.81 $~ &~$74.21\pm0.30$~ & ~$75.10\pm0.50$~   &  ~$78.45\pm0.26$~   & ~$55.52\pm0.46 $~&~$58.80\pm0.2$~       &~$71.26\pm1.04$~          &  ~$51.33\pm0.57$~  & ~$84.20\pm0.70$\\  \hline

  ~CORE SP~   & ~$88.29\pm1.55 $~ &~$73.46\pm0.32$~ & ~$-$~           &  ~$77.30\pm0.80$~   & ~$59.06\pm0.93 $~&~$-$~                 &~$72.62\pm0.59$~          &  ~$49.43\pm0.42$~  & ~$90.84\pm0.14$\\  \hline

  ~PIGK~   & ~$76.00\pm2.69 $~ &~$82.54\pm0.47$~ & ~$73.68\pm0.69$~   &  ~$78.25\pm0.51$~   & ~$59.50\pm2.44 $~ &~$-$~       &~$-$~          &  ~$-$~ & ~$-$\\ \hline

  ~  GK~   & ~$81.66\pm2.11 $~ &~$62.28\pm0.29$~ & ~$71.67\pm0.55$~   &  ~$78.45\pm0.26$~   & ~$52.26\pm1.41 $~ &~$72.83\pm0.28$~       &~$65.87\pm0.98$~          &  ~$45.42\pm0.87$~ & ~$77.34\pm0.18$\\ \hline

  ~RWGK~   & ~$80.77\pm0.72 $~ &~$63.34\pm0.27$~       &~$74.20\pm0.40$~          &  ~$71.70\pm0.47$~         & ~$55.91\pm0.37 $~  &~$-$~       &~$67.94\pm0.77$~          &  ~$46.72\pm0.30$~ & ~$-$\\ \hline



\end{tabular}
} \vspace{-10pt}
\end{table*}
\begin{table*}
\centering {
\tiny
\caption{Classification Accuracy (In $\%$ $\pm$ Standard Error) for Comparisons with Graph Convolutional Neural Networks.}\label{T:ClassificationGCNN}
\vspace{0pt}
\begin{tabular}{|c||c||c||c||c||c||c||c||c||c|}

  \hline
 ~Datasets~& ~MUTAG  ~       & ~NCI1~         & ~PROTEINS~      & ~D\&D~          & ~PTC(MR)~        & ~COLLAB~        & ~IBDM-B~        & ~IMDB-M~  & ~RED-B~    \\ \hline \hline

 ~\textbf{QSGCNN}~  & ~$\textbf{90.52}\pm0.95$~&~$77.50\pm0.91$~& ~$75.90\pm0.79$~& ~$\textbf{81.70}\pm0.92$~& ~$63.37\pm1.15$~ & ~$\textbf{78.80}\pm0.89$ ~& ~$\textbf{73.62}\pm1.12$ & ~$\textbf{51.60}\pm1.15$   & ~$\textbf{91.50}\pm0.24$\\ \hline

  ~DGCNN~  & ~$85.83\pm1.66$~&~$74.44\pm0.47$~& ~$75.54\pm0.94$~& ~$9.37\pm0.94$~& ~$58.59\pm2.47$~ & ~$73.76\pm0.49$ ~& ~$70.03\pm0.86$ & ~$47.83\pm0.85$   & ~$76.02\pm1.73$\\ \hline

  ~PSGCNN~   & ~$88.95\pm4.37$~&~$76.34\pm1.68$~& ~$75.00\pm2.51$~& ~$76.27\pm2.64$~& ~$62.29\pm5.68$~ & ~$72.60\pm2.15$ ~& ~$71.00\pm2.29$ & ~$45.23\pm2.84$  & ~$86.30\pm1.58$\\ \hline

  ~DCNN~   & ~$66.98$~       &~$56.61\pm1.04$~& ~$61.29\pm1.60$~& ~$58.09\pm0.53$~& ~$56.60$~   & ~$52.11\pm0.71$ ~& ~$49.06\pm1.37$ & ~$33.49\pm1.42$    & ~$-$\\ \hline

  ~ECC~    & ~$76.11$~       &~$76.82$~       & ~$72.65$~           & ~$74.10$~       & ~$-$~            & ~$67.79$~            & ~$-$            & ~$-$    & ~$-$\\ \hline


  ~GCCNN~& ~$-$~          &~$\textbf{82.72}\pm2.38$~            & ~$\textbf{76.40}\pm4.71$~           & ~$77.62\pm4.99$~       & ~$\textbf{66.01}\pm5.91$~            & ~$77.71\pm2.51$~            & ~$71.69\pm3.40$            & ~$48.50\pm4.10$   & ~$87.61\pm2.51$\\ \hline

  ~DGK~   & ~$82.66\pm1.45$~ &~$62.48\pm0.25$~& ~$71.68\pm0.50$~& ~$78.50\pm0.22$~    & ~$57.32\pm1.13$~ & ~$73.09\pm0.25$~& ~$66.96\pm0.56$ & ~$44.55\pm0.52$    & ~$78.30\pm0.30$\\ \hline


  ~AWE~& ~$87.87\pm9.76$~ &~$-$~          & ~$-$~           & ~$71.51\pm4.02$~    & ~$-$~          & ~$70.99\pm1.49$~& ~$73.13\pm3.28$ & ~$51.58\pm4.66$    & ~$82.97\pm2.86$\\ \hline



\end{tabular}
} \vspace{-10pt}
\end{table*}

\section{Experiments}\label{s4}
In this section, we empirically compare the performance of the proposed QSGCNN model to state-of-the-art graph kernels and deep learning methods on graph classification problems.

\subsection{Comparisons with Graph Kernels}

\noindent\textbf{Datasets:} In this subsection, we utilize nine standard graph datasets from bioinformatics~\cite{DBLP:journals/nar/SchomburgCEGHHS04,doi:10.1021/jm00106a046,bioinformatics2003,DBLP:journals/kais/WaleWK08} and social networks~\cite{socialnetworks2018} to evaluate the performance of the proposed QSGCNN model. These datasets include MUTAG, PTC, NCI1, PROTEINS, D\&D, COLLAB, IMDB-B, IMDB-M and RED-B. A selection of statistics of these datasets are shown in Table.\ref{T:GraphInformation}.\\

\noindent\textbf{Experimental Setup:} We evaluate the performance of the proposed QSGCNN model on graph classification problems against five alternative state-of-the-art graph kernels. These graph kernels include 1) Jensen-Tsallis q-difference kernel (JTQK) with $q=2$~\cite{DBLP:conf/pkdd/Bai0BH14}, 2) the Weisfeiler-Lehman subtree kernel (WLSK)~\cite{shervashidze2010weisfeiler}, 3) optima assignment Weisfeiler-Lehman kernel (WL-OA)~\cite{DBLP:conf/nips/KriegeGW16}, 4) the shortest path graph kernel (SPGK) \cite{DBLP:conf/icdm/BorgwardtK05}, 5) the shortest path kernel based on core variants (CORE SP)~\cite{DBLP:conf/ijcai/NikolentzosMLV18}, 6) the random walk graph kernel (RWGK)~\cite{DBLP:conf/icml/KashimaTI03}, 7) the graphlet count kernel (GK)~\cite{DBLP:journals/jmlr/ShervashidzeVPMB09}, and 8) the propagated information graph kernel (PIGK)~\cite{DBLP:journals/ml/NeumannGBK16}.

For the evaluation, \textbf{the proposed QSGCNN model uses the same network structure on all graph datasets}. Specifically, we set the number of the prototype representations as $M=64$, the number of the quantum spatial graph convolution layers as $5$ (note that, including the original input grid structures, the spatial graph convolution produces $6$ concatenated outputs), and the channels of each quantum spatial graph convolution as $32$. Following each of the concatenated outputs after the quantum graph convolution layers, we add a traditional CNN layer with the architecture as C$64$-P$2$-C$64$-P$2$-C$64$-F$64$ to learn the extracted patterns, where C$k$ denotes a traditional convolutional layer with $k$ channels, $P$k denotes a classical MaxPooling layer of size and stride $k$, and FC$k$ denotes a fully-connected layer consisting of $k$ hidden units. The filter size and stride of each C$k$ are all $5$ and $1$. With the six sets of extracted patterns after the CNN layers to hand, we concatenate them and add a new fully-connected layer followed by a Softmax layer with a dropout rate of $0.5$. We use the rectified linear units (ReLU) in either the graph convolution or the traditional convolution layer. The learning rate of the proposed model is $0.00005$ for all datasets. The only hyperparameter we optimized is the number of epochs and the batch size for the mini-batch gradient decent algorithm. To optimize the proposed QSGCNN model, we use the Stochastic Gradient Descent with the Adam updating rules. Finally, note that, the proposed QSGCNN model needs to construct the prototype representations to identify the transitive vertex alignment information over all graphs. The prototype representations can be computed from the training graphs or both the training and testing graphs. We observe that the proposed model associated with the two variants dose not influence the final performance. Thus, in our evaluation we proposed to compute the prototype representations from both the training and testing graphs. In this sense, our model can be seen as an instance of transductive learning~\cite{DBLP:conf/uai/GammermanAV98}, where all the graphs are used to compute the prototype representations, and the class labels of the test graphs are not observed during the training phase. For the proposed QSGCNN model, we perform $10$-fold cross-validation to compute the classification accuracies, with nine folds for training and one folds for testing. For each dataset, we repeat the experiment 10 times and report the average classification accuracies and standard errors in Table.\ref{T:ClassificationGK}.

We set the parameters controlling the maximum height of the subtrees for the Weisfeiler-Lehman isomorphism test (WLSK kernel) and for the tree-index method (JTQK kernel) to $10$. This is based on the previous empirical studies of Shervashidze et al.~\cite{shervashidze2010weisfeiler} and Bai et al.~\cite{DBLP:conf/pkdd/Bai0BH14}. For each graph kernel, we perform $10$-fold cross-validation using the LIBSVM implementation of C-Support Vector Machines (C-SVM) and we compute the classification accuracies. We perform cross-validation on the training data to select the optimal parameters for each kernel and fold. We repeat the experiment 10 times for each kernel and dataset and we report the average classification accuracies and standard errors in Table.\ref{T:ClassificationGK}. Note that for some kernels we directly report the best results from the original corresponding papers, since the evaluation of these kernels followed the same setting of ours.\\

\noindent\textbf{Experimental Results and Discussion:} Table.\ref{T:ClassificationGK} shows that the proposed QSGCNN model significantly outperforms the alternative state-of-the-art graph kernels in this study. Although, the proposed model cannot achieve the best classification accuracy on the NCI1 and COLLAB datasets, but the proposed model is still competitive and the accuracy on the COLLAB dataset is only a little lower than the WL-OA kernel. On the other hand, the accuracy of the proposed model on the NCI1 dataset is still higher than the SPGK, CORE SP, GK and RWGK kernels. The reasons for the effectiveness are twofold. First, the state-of-the-art graph kernels for comparisons are typical examples of R-convolution kernels. Specifically, these kernels are based on the isomorphism measure between any pair of substructures, ignoring the structure correspondence information between the substructures. By contrast, the associated aligned vertex grid structure for the proposed QSGCNN model incorporates the transitive alignment information between vertex over all graphs. Thus, the proposed model can better reflect the precise characteristics of graphs. Second, the C-SVM classifier associated with graph kernels can only be seen as a shallow learning framework~\cite{DBLP:conf/icassp/ZhangLYG15}. By contrast, the proposed QSGCNN model can provide an end-to-end deep learning architecture for graph classification, and can better learn the graph characteristics. The experiments demonstrate the advantages of the proposed QSGCNN model, compared to the shallow learning framework. Third, some alternative kernels are related to the Weisfeiler-Lehman method. As we have stated in Section~\ref{s3.4}, the kernels based on the Weisfeiler-Lehman method may suffer from the tottering problem. By contrast, the proposed model based on quantum walk can significantly reduce the effect of tottering walks. The experiments also demonstrate the effectiveness.

\subsection{Comparisons with Deep Learning Mthods}
\noindent\textbf{Datasets:} In this subsection, we further compare the performance of the proposed QSGCNN model with state-of-the-art deep learning methods for graph classifications. The datasets for the evaluations include the mentioned five datsets from bioinformatics, as well as three social network datasets. The social network datasets include COLLAB, IMDB-B, and IMDB-M. Details of these social network datasets can be found in Table.\ref{T:GraphInformation}.\\

\noindent\textbf{Experimental Setup:} We evaluate the performance of the proposed QSGCNN model on graph classification problems against five alternative state-of-the-art deep learning methods for graphs. These methods include 1) the deep graph convolutional neural network (DGCNN)~\cite{DBLP:conf/aaai/ZhangCNC18}, 2) the PATCHY-SAN based convolutional neural network for graphs (PSGCNN)~\cite{DBLP:conf/icml/NiepertAK16}, 3) the diffusion convolutional neural network (DCNN)~\cite{DBLP:conf/nips/AtwoodT16}, 4) the edge-conditioned convolutional networks (ECC)~\cite{DBLP:conf/cvpr/SimonovskyK17}, 5) the deep graphlet kernel (DGK)~\cite{DBLP:conf/kdd/YanardagV15}, 6) the graph capsule convolutional neural network (GCCNN)~\cite{DBLP:journals/corr/abs-1805-08090}, and 7) the anonymous walk embeddings based on feature driven (AWE)~\cite{DBLP:conf/icml/IvanovB18}. For the proposed QSGCNN model, we use the same experimental setups when we compare the proposed model to graph kernels. For the PSGCNN, ECC, and DGK model, we report the best results from the original papers~\cite{DBLP:conf/icml/NiepertAK16,DBLP:conf/cvpr/SimonovskyK17,DBLP:conf/kdd/YanardagV15}. Note that, these methods follow the same setting with the proposed QSGCNN model. For the DCNN model, we report the best results from the work of Zhang et al.,~\cite{DBLP:conf/aaai/ZhangCNC18}, following the same setting of ours. For the AWE model, we report the classification accuracies of the feature-driven AWE, since the author have stated that this kind of AWE model can achieve competitive performance on label dataset. Finally, the PSCN and ECC models can leverage additional edge features. Since most graph datasets and all the alternative methods to used for comparisons do not leverage edge features, in this work we do not report the results associated with edge features. The classification accuracies and standard errors for each deep learning method are shown in Table.\ref{T:ClassificationGCNN}.\\

\noindent\textbf{Experimental Results and Discussion:} Table~\ref{T:ClassificationGCNN} indicates that the proposed QSGCNN model significantly outperforms state-of-the-art deep learning methods for graph classifications, on the MUTAG, D\&D, COLLAB, IBDM-B, IBDM-M and RET-B datasets. On the other hand, only the accuracy of the GCCNN model on the NCI1 and PTC datasets and that of the DGCNN model on the PROTEINS dataset are a higher than the proposed QSGCNN model. But the proposed QSGCNN is still competitive and outperform the remaining methods on the three datasets. The reasons of the effectiveness are fivefold.

First, similar to the state-of-the-art graph kernels, all the alternative deep learning methods (i.e., the DGCNN, PSGCNN, DCNN, ECC, GCCNN, DGK and AWE models) for comparisons also cannot integrate the correspondence information between graphs into the learning architecture. Especially, the PSGCNN, DGCNN and ECC models need to reorder the vertices, but these methods rely on simple but inaccurate heuristics to align the vertices of the graphs, i.e., they sort the vertex orders based on the local structure descriptor of each individual graph and ignore the vertex correspondence information between different graphs. Thus, only the proposed QSDCNN model can precisely reflect the graph characteristics through the layer-wise learning.

Second, the PSGCNN and DGCNN models need to form a fixed-sized vertex grid structure for each graph. Since the vertex numbers of different graphs are different, forming such sixed-sized grid structures means some vertices of each graph may be discarded, leading to information loss. By contrast, as we have mentioned in Section~\ref{s2} and Section~\ref{s3}, the associated aligned vertex grid structures can completely preserve the information of original graphs. As a result, only the proposed QSGCNN model can completely integrate the original graph characteristics into the learning process.

Third, unlike the proposed model, the DCNN model needs to sum up the extracted local-level vertex features from the convolution operation as global-level graph features through a SumPooling layer. Thus, only the QSGCNN model can learn the graph topological information through the local vertex features.

Forth, unlike the PSGCNN, DGCNN, GCCNN and ECC models that are based on the original vertex adjacency matrix to formulate vertex connection information of the graph convolution operation, the graph convolution operation of the proposed QSGCNN model formulates the vertex connection information in terms of the average mixing matrix of continuous-time quantum walk. As we have stated in Section~\ref{s2}, the quantum walk is not dominated by the low frequency of the Laplacian spectrum and can better distinguish different graph structures. Thus, the proposed QSDCNN model has better ability to identify the difference between different graphs.

Fifth, similar to the DGCNN, PSGCNN and DGK models, the proposed QSGCNN model is also related to the classical Weisfeiler-Lehman (WL) method. Since the classical WL method suffers from tottering problem, the related DGCNN, PSGCNN and DGK models also process the same drawback. By contrast, the graph convolution operation of the proposed QSGCNN model can be seen as the quantum version of the classical WL algorithm. Since the quantum walk can reduce the problem of tottering problem, the proposed QSGCNN model overcomes the shortcoming of tottering problem arising in the DGCNN, PSGCNN and DGK models. Sixth, the AWE model is based on the classical random walk. By contrast, the proposed QSGCNN model is based on the quantum random walk, that has been proven powerful to better distinguish different graph structures. The evaluation demonstrates the advantages of the proposed QSGCNN model, compared to the state-of-the-art deep learning methods.

\section{Conclusion}\label{s6}
In this paper we have developed a new Quantum Spatial Graph Convolutional Neural Network (QSGCNN) model, that can directly learn an end-to-end deep learning architecture for classifying graphs of arbitrary sizes. The main idea of the proposed QSGCNN model is to transform each graph into a fixed-sized vertex grid structure through transitive alignment between graphs and propagate the grid vertex features using the proposed quantum spatial graph convolution operation. Compared to state-of-the-are deep learning methods and graph kernels, the proposed QSGCNN model cannot only preserve the original graph characteristics, but also bridge the gap between the spatial graph convolution layer and the traditional convolutional neural network layer. Moreover, the proposed QSGCNN can better distinguish different structures, and the experimental evaluations demonstrate the effectiveness of the proposed QSGCNN model on graph classification problems.

In this work, we used the same network architecture for all datasets. In future works, we aim to learning the optimal structure for each dataset, which in turn should lead to improved performance. Furthermore, in future works, we also aim to extend the proposed QSGCNN model and develop a new quantum graph neural network drawing on edge-based grid structures. In previous works~\cite{DBLP:conf/iciap/BaiZR0H15,DBLP:journals/prl/BaiRCZRBH17,DBLP:journals/pr/BaiEH16} we have shown how to characterize the edge information of the original graphs through the directed line graphs, where each vertex of the line graph represents an edge of original graphs. Moreover, we have illustrated the relationship between the discrete-time quantum walks and the directed line graphs. It will be interesting to develop a novel quantum edge-based convolutional network associated with the discrete-time quantum walks and the directed line graphs.

Finally, note that, Xu et al.,~\cite{DBLP:journals/corr/abs-1810-00826} have recently indicated that the convolutional operation of most existing graph convolutional neural networks associated with the adjacency matrix can be seen as directly employing a $1$-layer perceptron followed by a non-linear activation function such as a ReLU. Moreover, they developed a new graph isomorphism network model based on a new vertex information aggregation layer followed by multi layer perceptrons. They demonstrated that this can significantly improve the performance of state-of-the-art graph convolutional networks. This work enlightens our future work, and we will further extend the proposed QSGCNN model into a new quantum isomorphism network.

\section*{Acknowledgments}
This work is supported by the National Natural Science Foundation of China (Grant no.61503422 and 61602535), the Open Projects Program of National Laboratory of Pattern Recognition (NLPR), and the program for innovation research in Central University of Finance and Economics.

\balance


\bibliographystyle{IEEEtran}
\bibliography{example_paper}

\end{document}